%% file: main.tex
\documentclass{article}
\PassOptionsToPackage{table}{xcolor}
\usepackage{iclr2027_conference,times}

\iclrfinalcopy

\usepackage{hyperref}
\usepackage{url}
\usepackage{graphicx}
\usepackage{wrapfig}
\usepackage{booktabs}
\usepackage{amsmath,amssymb}
\usepackage{array}
\usepackage{multirow}
\usepackage[table]{xcolor}
\usepackage{subcaption}
\usepackage[capitalize,noabbrev]{cleveref}

\graphicspath{{figs/}{../figs/}}

\definecolor{goodcell}{HTML}{DCEFE0}
\definecolor{badcell}{HTML}{F7DCDA}
\newcommand{\good}[1]{\cellcolor{goodcell}#1}
\newcommand{\bad}[1]{\cellcolor{badcell}#1}

\definecolor{cpaired}{HTML}{D1372F}
\definecolor{cstepdown}{HTML}{E07B39}
\definecolor{cstepup}{HTML}{2A78D6}
\definecolor{clate30}{HTML}{1BAF7A}

\newcommand{\LU}{LUC\ensuremath{_{0.5}}}
\newcommand{\LUd}{LUC\ensuremath{_{0.2}}}
\newcommand{\EU}{EUC\ensuremath{_{0.5}}}
\newcommand{\EUd}{EUC\ensuremath{_{0.2}}}

\newcommand{\mpaired}{\textcolor{cpaired}{\textbf{paired}}}
\newcommand{\mLU}{\textcolor{cstepup}{\textbf{LUC}\ensuremath{_{0.5}}}}
\newcommand{\mLUd}{\textcolor{clate30}{\textbf{LUC}\ensuremath{_{0.2}}}}

\title{When to Unpair: Regulating Pairing Dependence in Medical Visual In-Context Learning}

\author{%
Cheng Wan, Chenjun Li \& Qingyu Zhao\\
Cornell University\\
Weill Cornell Medicine\\
\texttt{\{cw2222, cl2733, qz387\}@cornell.edu}
}

\begin{document}
\maketitle

\fancyhead{}

\input{sections/body}

\bibliographystyle{iclr2027_conference}
\bibliography{iclr2027_conference}

\clearpage
\appendix
\input{sections/appendix}

\end{document}

%% file: sections/body.tex
\begin{abstract}
Visual in-context learning (ICL), well suited to label-scarce medical imaging, uses support image--label pairs to demonstrate input--output mappings, while the labels collectively indicate the requested task.
We diagnose dependence on individual pairings with a test-time \emph{derangement} that reassigns every support label to another support image while preserving the query, support images, and label multiset.
The resulting \emph{pairing gap}, defined as shuffled-minus-matched performance, shows that all four released models depend on the pairing, to widely varying degrees.
Further analysis of a paired-trained model reveals support-associated spurious regions and lesion-size biases even with real, unaltered supports, alongside sensitivity to mis-registered support labels.
To regulate this dependence, we introduce a \emph{late unpairing curriculum} (LUC), which starts with matched training and then applies \emph{random unpairing}, replacing each support label with that of another support in the same episode.
LUC nearly closes the pairing gap on two backbones while maintaining or improving matched-support performance across all evaluated task types, with gains extending to held-out tasks and cross-dataset episodes.
It also mitigates these failure modes.
On BraTS whole-tumor segmentation, matched-support DSC rises from $0.733$ to $0.857$ while the gap shrinks from $-0.184$ to $-0.008$.
In a released model, brief fine-tuning with random unpairing reduces the gap.
A reversed curriculum that places the same number of unpairing epochs at the start of training leaves a large gap.
This shows that pairing dependence is shaped by the order of training and not only by the amount of unpaired training.
\end{abstract}

\section{Introduction}\label{sec:intro}
\begin{figure}[h]\centering
\includegraphics[width=\textwidth]{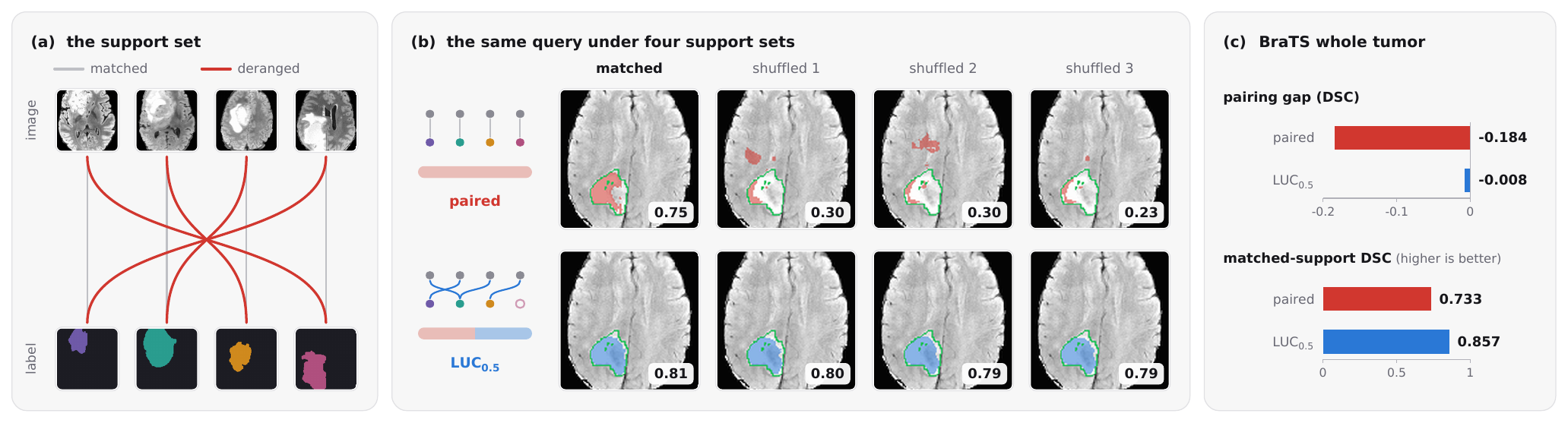}
\caption{\textbf{Which label sits on which support image changes the prediction.}
(a) Grey: every support image keeps its own label; red: a \emph{derangement}, which moves each label onto another image and leaves the label multiset unchanged.
(b) One BraTS validation case at $K{=}4$ with matched supports and three derangements; beside each row are the support set that model trains on and its schedule $p(e)$. Green outlines the ground truth; numbers are DSC.
(c) Pairing gap and matched-support DSC over the BraTS whole-tumor evaluation.}
\label{fig:teaser}
\end{figure}
In-context learning (ICL) lets a single model solve a new task from a few examples, with no task-specific retraining~\citep{brown2020language}.
In its visual form the model receives a query image and a few support (image, label) pairs and returns a dense prediction, such as a segmentation or an image in another modality~\citep{bar2022visual, wang2023images, wang2023seggpt, kim2023universal}.
This suits medical imaging, where dense labels are scarce and expensive, and medical ICL models now cover 2D and 3D segmentation and image-to-image tasks~\citep{butoi2023universeg, czolbe2023neuralizer, rakic2024tyche, hu2024neuroverse3d, hu2026medverse}.

An underexplored question in ICL is how models use the supports to infer the task.
Support pairs provide task-relevant information in complementary ways.
A model can use the image--label correspondence within individual support pairs to infer how inputs map to outputs, while the collection of support labels can indicate what output is requested.
For example, support masks can collectively distinguish whole-tumor from edema segmentation, while their pairing with individual images provides example-specific mappings between image appearance and label.
Both sources can be informative, but how current visual ICL models balance them, and how strongly their predictions depend on individual image--label pairings, remains poorly understood.

We study this behavior through a test-time derangement that selectively disrupts individual image--label correspondence.
Specifically, we preserve the query image, all support images, and the same set of support labels, but reassign each label to a different support image, which gives shuffled supports (\cref{fig:teaser}).
Comparing performance under matched and shuffled supports yields a pairing gap, which quantifies a model's dependence on individual support pairings while holding the contents of the support set fixed.
In doing so, we find substantial variation in pairing dependence across existing visual ICL models (\cref{tab:generic-shortcut}) and show that strong dependence can have undesirable consequences.
In medical image segmentation, for example, the predicted lesion extent and location can be influenced by the size and spatial position of lesions in the support labels (\cref{sec:failure}).

Building on this diagnostic, we find that training-time random unpairing can regulate how models use support pairing.
Replacing support labels during training with those of other supports in the same episode weakens the reliability of individual image--label correspondence, while the labels, taken together, still indicate which output is requested.
The timing of this intervention matters: a late unpairing curriculum (LUC), which unpairs only after an initial period of conventional paired training, substantially reduces test-time pairing dependence.
This robustness comes without a trade-off in standard predictive performance: LUC maintains or improves performance under correctly matched supports across all evaluated task types.
It also mitigates several failure modes associated with excessive dependence on individual support pairs.
Together, the test-time derangement and training-time unpairing form a strategy for diagnosing and controlling pairing dependence in visual ICL: the former reveals how a trained model uses support correspondence, and the latter reshapes that behavior without changing the model architecture or giving up matched-support performance.

Our contributions are:
\begin{itemize}\itemsep1pt
  \item A test-time derangement of the support labels as a black-box diagnostic of pairing dependence, and the finding that this dependence varies widely across four released visual in-context learners (\cref{sec:shortcut}).
  \item The failure modes of strong dependence in paired training: the prediction follows the support labels rather than the query, with real support sets and under controlled perturbations of the labels, and loses accuracy when the support labels are mis-registered (\cref{sec:failure}).
  \item LUC, a schedule for random unpairing that substantially reduces pairing dependence while maintaining or improving matched-support performance on five task types and two backbones, and that also reduces these failure modes (\cref{sec:both}). The timing matters: with the same number of unpairing epochs, the schedule that ends on matched supports keeps a large gap. On a released model, $1000$ fine-tuning steps of random unpairing shrink the edema gap from $-0.353$ to $-0.036$.
\end{itemize}

\section{Related work}\label{sec:related}
\paragraph{In-context learning.}
In-context learning lets a model solve a task from a few input--output examples placed in its input~\citep{brown2020language}.
In language models, how much the example labels matter depends on the setting: random labels barely lower accuracy on classification tasks~\citep{min2022rethinking}, while larger models follow flipped labels~\citep{wei2023larger} and ICL does learn the label relationships in its context~\citep{kossen2024incontext}.
Recognizing which task the examples name can be separated from learning the task in context~\citep{pan2023context}, transformers can solve ICL by copying from matching context~\citep{olsson2022context}, and the training data decide whether ICL emerges and whether it persists~\citep{chan2022data, singh2023transient}.
We study pairing dependence in visual ICL and how it changes when support labels are reassigned during training.

\paragraph{Visual and medical in-context learning.}
Visual ICL defines a dense prediction task through a few (image, label) examples, first by inpainting a canvas that holds the examples and the query~\citep{bar2022visual, wang2023images} and then at scale and for segmentation and other dense tasks~\citep{bai2024sequential, wang2023seggpt, kim2023universal}.
Segment Anything~\citep{kirillov2023segment} can also be adapted to segment a target from one reference image and its mask~\citep{zhang2024personalize, liu2024matcher}, and which examples are shown changes the result~\citep{zhang2023makes}.
In medical imaging, where labels are scarce, few-shot segmentation~\citep{ouyang2020self} has grown into ICL models that solve unseen tasks from a support set without retraining: UniverSeg and Tyche in 2D~\citep{butoi2023universeg, rakic2024tyche}, and Neuralizer, Neuroverse3D, Medverse and Iris for multi-task and 3D imaging~\citep{czolbe2023neuralizer, hu2024neuroverse3d, hu2026medverse, gao2025show}.
Universal models that take a spatial or text prompt instead of examples form a parallel line~\citep{ma2024segment, wang2025sam, liu2023clip}.
Support labels are drawn by hand and vary within and between observers~\citep{karimi2020deep}, and in-context segmenters lose accuracy when their support labels are deformed~\citep{sheng2025unicl}.
The ICL models are trained and evaluated with matched supports; we measure what four released models do when the pairing changes, and we change the pairing during training.

\paragraph{Training on support sets.}
Episodic training builds each step from a support set and a query~\citep{vinyals2016matching, snell2017prototypical}.
When a query can be solved without its supports, a meta-learner can learn to ignore them, which a random assignment of labels per task prevents~\citep{yin2020meta, rajendran2020meta}.
Random unpairing acts inside one support set and targets the opposite failure, excessive reliance on individual pairings: while it is applied, the labels still indicate the task as a set, but which image each label sits on carries no information.

\paragraph{Curriculum learning.}
Curriculum learning orders the training signal, classically from easy to hard examples~\citep{bengio2009curriculum}, and later work chooses the order automatically and studies when it helps~\citep{graves2017automated, hacohen2019power, wu2021curricula, soviany2022curriculum}.
The order also shapes what a network keeps: deficits early in training can have lasting effects~\citep{achille2018critical}.
LUC orders how reliable the support pairs are rather than how hard the examples are, starting on matched supports and ending on unpaired ones; for the pairing gap we find the reverse of a lasting early effect, since the phase that ends training sets the gap.

\section{Method}\label{sec:method}
\begin{figure}[t]\centering
\includegraphics[width=\textwidth]{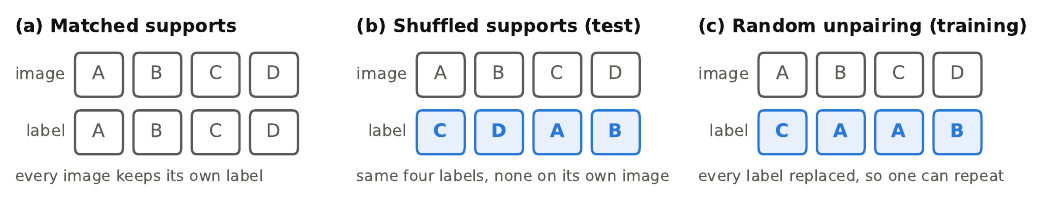}
\caption{\textbf{The two operations on the support set.}
(a) Matched supports.
(b) The test-time derangement: the same four labels, none on its own image.
(c) Random unpairing in training: every support takes a copy of another support's label, so a label can repeat, which never happens in (b).}
\label{fig:protocol}
\end{figure}

We measure a model's dependence on support pairings with a test-time derangement and then propose training-time random unpairing to regulate this dependence.
\Cref{fig:protocol} shows both operations on a support set of four, and \cref{app:impl} gives the full protocol.
We describe the method in the context of Neuroverse3D~\citep{hu2024neuroverse3d}, but the same formulation applies to all models tested here.

\paragraph{The visual in-context forward pass.}
Given a query $x_q$ and a support set $S=\{(x_i,y_i)\}_{i=1}^{K}$, Neuroverse3D uses a shared-weight support branch $g$ to process each pair $(x_i,y_i)$ and predicts a dense label $\hat{y}_q$ with a target encoder $f_{\text{enc}}$ for the query.
Support representations are combined by an equal-weight mean at every decoder stage $s$,
\begin{equation}
  \bar{c}^{\,s} \;=\; \frac{1}{K}\sum_{i=1}^{K} g_s\!\bigl(x_i\oplus y_i,\; f^s_{\text{enc}}(x_q)\bigr),
  \label{eq:agg}
\end{equation}
where $\oplus$ is channel concatenation, and a target decoder fuses $\bar{c}^{\,s}$ with skip connections to produce $\hat{y}_q$ ($\sim$71M parameters).
The mean in \cref{eq:agg} is invariant to reordering whole support pairs but not to permuting their labels, since each image and its assigned label interact inside $g_s$ before aggregation.
Any dependence on the matching therefore comes from the training distribution rather than from the aggregation.

\paragraph{Test-time derangement of the support labels.}
To measure how strongly a trained model depends on the assignment, we replace the matched supports with shuffled ones, $S_{\text{shuf}}=\{(x_i,\,y_{\sigma(i)})\}_{i=1}^{K}$, where $\sigma$ is a \emph{derangement}, a permutation that leaves no label on its own image and exists for $K\ge 2$, drawn by rejection sampling.
Every support label now sits on another support image, and the label multiset of the support set is unchanged.
We summarize pairing dependence by the pairing gap
\begin{equation}
  \Delta = M_{\text{shuffled}}-M_{\text{matched}},
  \label{eq:gap}
\end{equation}
where $M$ is the task metric (DSC or PSNR) and the support set is resampled per query; a model that depends on the pairing loses accuracy under the derangement and has a negative gap.

\paragraph{Late unpairing curriculum.}
We regulate pairing dependence in training with random unpairing.
In an unpairing epoch, every support in every episode has its label replaced by a copy of the label of another support in the same set, chosen at random; the query and its target are unchanged.
A label can therefore appear twice in the support set while another is dropped, so the training operator does not preserve the label multiset, while the test-time derangement does.
The operation is the same for every model we train; what differs is the schedule $p(e)$, which marks each epoch $e$ as matched ($p(e){=}0$) or unpairing ($p(e){=}1$).
The \emph{late unpairing curriculum} (LUC) keeps $p(e){=}0$ for the first $e_0$ epochs and sets $p(e){=}1$ afterwards, so the model first learns from matched supports and then finishes training on unpaired ones.
We write LUC$_{x}$, where $x$ is the fraction of training epochs with unpairing: \LU{} unpairs in the second half of training, and \LUd{} only in the final fifth.
\LU{} is our primary schedule, and LUC without a subscript refers to it when results are reported.
The \emph{early unpairing curriculum} (EUC$_{x}$) runs the same schedule in reverse, unpairing first and ending on matched supports, and serves as a control for timing, since it unpairs for as many epochs as LUC$_{x}$.
Two constant schedules complete the family: $p(e){=}0$ throughout, which is conventional paired training, and $p(e){=}1$ throughout, the \emph{unpaired} model.
Every Neuroverse3D model we train uses one recipe, detailed in \cref{app:impl}, on a mix of five task types from BraTS and three T1 brain MRI cohorts: tumor segmentation, anatomical segmentation, modality translation, inpainting and bias correction.

\section{Pairing dependence and the failure modes of paired training}\label{sec:shortcut}
\textbf{Pairing gap of ICL models.} The derangement is a black-box, inference-only test, so we apply it to four released models to derive their pairing gap (\cref{tab:generic-shortcut}; per-model protocols in \cref{app:impl}).
On the two BraTS segmentation tasks~\citep{menze2014multimodal}, the four models differ substantially in how strongly they depend on the pairing.
On the whole-tumor segmentation task, the test-time derangement induces a DSC drop of at least $0.11$ (for Neuroverse3D), with the largest drop, $0.63$, observed for the general-purpose SegGPT.
For the edema segmentation task, the pairing gap ranges from $-0.35$ to $-0.51$.
The gap also differs between tasks for the same model; Neuroverse3D, for example, has a gap of $-0.11$ on whole-tumor segmentation and $-0.35$ on edema segmentation.

\begin{table}[t]\centering
\caption{\textbf{Pairing gap in four released models on two BraTS tasks.}
Each model is run with its own protocol using released weights (inference only); we report DSC on matched supports and the pairing gap, defined as the difference between shuffled and matched DSC.}
\label{tab:generic-shortcut}
{\footnotesize\setlength{\tabcolsep}{4.5pt}
\begin{tabular}{@{}ll cc cc@{}}
\toprule
 & & \multicolumn{2}{c}{Whole tumor} & \multicolumn{2}{c}{Edema} \\
\cmidrule(lr){3-4}\cmidrule(lr){5-6}
Backbone & Fusion & matched & gap & matched & gap \\
\midrule
\multicolumn{6}{@{}l}{\textit{2D}}\\
UniverSeg~\citep{butoi2023universeg}      & averaging        & 0.731 & $-0.41$          & 0.508 & $-0.36$ \\
SegGPT~\citep{wang2023seggpt}             & feature ensemble & 0.713 & $-0.63$          & 0.510 & $-0.51$ \\
\addlinespace[2pt]
\multicolumn{6}{@{}l}{\textit{3D}}\\
Neuroverse3D~\citep{hu2024neuroverse3d}   & mean-over-$K$    & 0.846 & $-0.11$          & 0.593 & $-0.35$ \\
Medverse~\citep{hu2026medverse}                 & cross-attention  & 0.824 & $-0.25$          & 0.654 & $-0.36$ \\
\bottomrule
\end{tabular}\par}
\end{table}

\begin{figure}[t]\centering
\includegraphics[width=\textwidth]{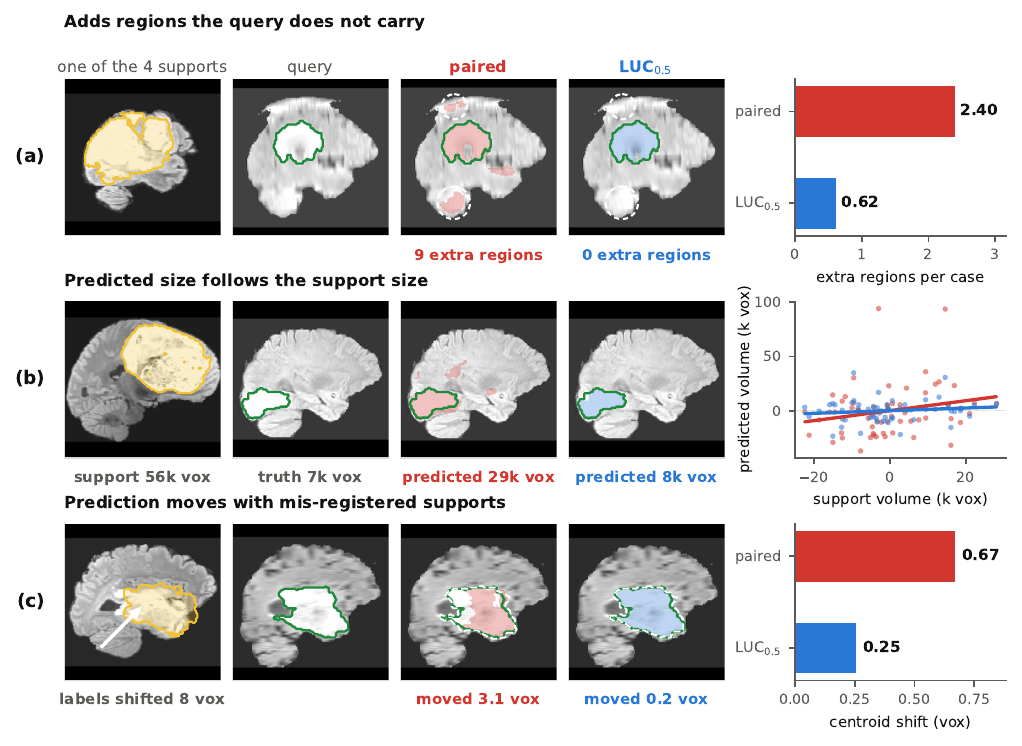}
\caption{\textbf{The influence of support labels on query predictions.}
%(a, b) Real support sets with nothing altered; (c) a controlled perturbation of the support labels.
Green outline: ground truth label; filled: model's prediction.
The first column shows one of the $K{=}4$ support images, with its label in yellow.
Each row shows a prediction failure, with the right-hand panel reporting statistics of the artifact over $n{=}60$ validation subjects.
(a) Based on real, unaltered supports, the paired model predicts additional regions that do not overlap the ground truth (white dashed circles).
(b) Based on real, unaltered supports, volumes of the tumor predicted by the paired model correlate with tumor size in the support. \LU{} removes this spurious correlation.
%are shown for the support labels, predictions, and ground truth; in the right-hand panel both volumes are shown after removing the part explained by the true tumor volume.
(c) Shifting every support label by $8$ voxels along each of two axes (white arrow in the first column) causes the paired model's prediction to shift in the same direction;
%Every support label is shifted by $8$ voxels along each of two axes; 
in the prediction panels, the dashed outline is the prediction with the original supports and the filled region the prediction with the shifted supports.}
\label{fig:failure-modes}
\end{figure}

\textbf{Support-label properties influence query predictions.} The pairing gap measures how much a prediction depends on which label sits on which image; it does not show what this dependence does to the prediction.\label{sec:failure}
Under matched supports, each support label is the correct answer for its own image, so a model can take parts of its answer, such as where the tumor lies and how large it is, from the support labels instead of from the query.
%We call this copying the support labels.
We examine whether such support-specific properties carry over to the query prediction in the Neuroverse3D model trained with matched supports only ($p{=}0$), the \emph{paired model}.
%that we trained with matched supports only ($p{=}0$), the \emph{paired model}, 
Using $60$ BraTS validation subjects with $K{=}4$ supports, we first inspect this behavior with real, unaltered support sets and then test it directly through controlled perturbations of the support labels (\cref{fig:failure-modes}).

With real, unaltered support sets, we observe that the model tends to segment regions that are absent in the query image.
On average, the model generates $2.40$ such extra regions per case (\cref{fig:failure-modes}a).
These regions are not randomly distributed: they tend to lie where the support labels carry tumor, and simply deleting these extra regions from the segmentation would raise the DSC by $0.041$, which indicates that the support-associated behavior accounts for a measurable fraction of the model's error.
Moreover, we observe that the predicted tumor size also follows the support labels: after accounting for the true tumor volume, larger support labels are associated with larger predicted tumor volumes (\cref{fig:failure-modes}b).
\Cref{app:failure} gives details on counting the extra regions and on the volume analysis.

\textbf{Controlled perturbations of support-label properties.} 
Given the above evidence, we next causally test whether properties of the support labels influence the query prediction. We perturb the support labels while leaving the query and support images unchanged, using two interventions.
%The derangement is one controlled perturbation of the support labels: it leaves the query untouched, so a prediction read from the query should not move.
%Two more perturbations of this kind test this carry-over directly.
First, we shift every support label by $8$ voxels along each of two axes, as a mis-registration would.
The model's predicted tumor centroid moves $0.67$ voxels along the shift (\cref{fig:failure-modes}c). This perturbation also reduces whole-tumor DSC by $0.035$ (\cref{fig:safety} in \cref{app:results} shows this cost next to replaced and eroded support labels), showing that the transferred spatial bias has a measurable effect on segmentation accuracy.
Next, we inject one synthetic mask into every support label, at a place where the query has no tumor.  This intervention makes the model predict most of the injected region as tumor: the fraction of mask voxels predicted as tumor rises by $0.846$ (\cref{fig:leakage} in \cref{app:failure} shows an example).
%In practice support labels are drawn by hand, so a label can end up with another patient's scan or slightly out of register, and both errors cost the paired model accuracy.
%As $0$ to $4$ of its four support labels are replaced by other patients' labels, its whole-tumor DSC falls from $0.74$ to $0.62$, and the $8$-voxel shift costs it $0.035$ DSC (\cref{fig:safety} in \cref{app:results} plots both, together with eroded labels).
In every one of these tests, the prediction of the model follows the support labels rather than the query.

\section{LUC reduces the dependence and the failure modes}\label{sec:both}
Paired training gives a model no reason to doubt a single support label, because every support label belongs to its own image.
We want a model that still learns the task from its supports but no longer takes its answer from individual support labels.
Random unpairing (\cref{sec:method}) supplies that reason: once each support carries another support's label, a single label is no longer a reliable answer, while the labels as a set still show which output is requested.
LUC applies it late, after a phase of matched training in which the model learns from correct pairs how images map to labels.

\subsection{Pairing gap and matched-support performance}\label{sec:gap-effect}

\begin{figure}[t]\centering
\includegraphics[width=\textwidth]{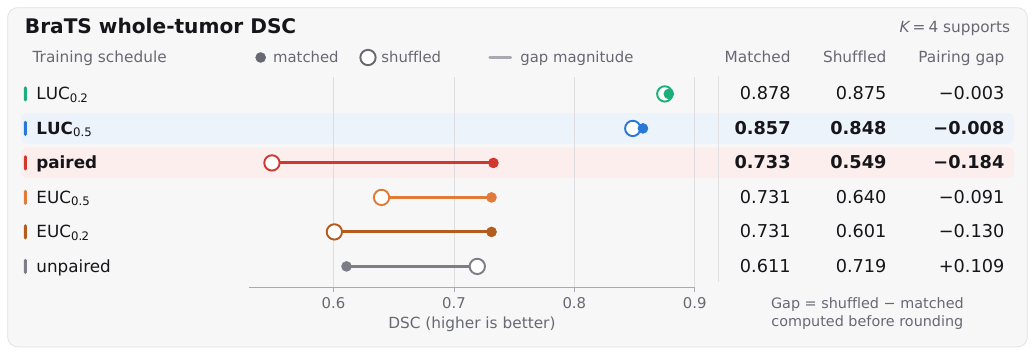}
\caption{\textbf{Matched-support and shuffled-support DSC on BraTS whole-tumor segmentation.}
Filled marker: DSC evaluated on matched supports; open marker: DSC evaluated with the shuffled support labels (test-time derangement); the line between them is the pairing gap.
LUC combines a near-zero pairing gap with high matched-support performance. Reversing the schedule with EUC retains a larger gap, while unpairing in every epoch (the unpaired model) lowers matched-support performance.}
%The paired model loses the most, \EU{} keeps a long line, and both LUC models sit furthest right with almost no line.
%Raising a constant unpairing probability also shortens the line, at a lower matched-support score.}
\label{fig:both}
\end{figure}

\textbf{LUC closes the pairing gap.} Under the test-time derangement, LUC predictions barely depend on which label sits on which support image.
On the BraTS whole-tumor segmentation task, the gap shrinks from $-0.184$ for the paired model to $-0.008$ for \LU{} (\cref{fig:both}), and on all five trained task types it is near zero for both LUC models (\cref{tab:both}).
The reduction extends beyond BraTS and beyond one backbone.
On three anatomical cohorts spanning children in ABCD~\citep{casey2018adolescent}, older adults in ADNI~\citep{jack2008adni} and a mixed population in OASIS-3~\citep{lamontagne2019oasis}, \LU{} leaves a gap within $0.004$ DSC of zero (\cref{tab:app-pop} in \cref{app:results} gives the gap for each cohort). When applied to 2D UniverSeg trained from scratch on BraTS multi-region segmentation,
%where the supports specify which region to segment, 
LUC reduces the gap from $-0.222$ to about zero (\cref{tab:generic-fix} in \cref{app:results} compares the two backbones).

\begin{table}[t]\centering
\caption{\textbf{Matched-support score and pairing gap on five task types.}
The models differ only in the unpairing schedule $p(e)$.
\good{Green}: an LUC cell with a near-zero gap and a matched-support score at least equal to that of the paired model; \bad{red}: the paired model's gap.
Per-task detail is in \cref{app:results}.}
\label{tab:both}\small
\setlength{\tabcolsep}{5pt}
\begin{tabular}{l cc cc cc}
\toprule
 & \multicolumn{2}{c}{\mpaired\ ($p{=}0$)} & \multicolumn{2}{c}{\mLU} & \multicolumn{2}{c}{\mLUd} \\
\cmidrule(lr){2-3}\cmidrule(lr){4-5}\cmidrule(lr){6-7}
\textbf{Task (metric)} & matched & gap & matched & gap & matched & gap \\
\midrule
Segmentation (DSC\,$\uparrow$)     & $0.733$ & \bad{$-0.184$} & \good{$0.857$} & \good{$-0.008$} & \good{$0.878$} & \good{$-0.003$} \\
Translation (PSNR\,$\uparrow$)     & $22.41$ & \bad{$-1.22$}  & \good{$22.56$} & \good{$-0.03$}  & \good{$22.70$} & \good{$-0.08$}  \\
Anatomical seg.\ (DSC\,$\uparrow$) & $0.704$ & \bad{$-0.035$} & \good{$0.806$} & \good{$-0.001$} & \good{$0.791$} & \good{$-0.007$} \\
Inpainting (PSNR\,$\uparrow$)      & $17.52$ & \bad{$-2.27$}  & \good{$17.76$} & \good{$+0.05$}  & \good{$17.88$} & \good{$-0.02$}  \\
Bias corr.\ (PSNR\,$\uparrow$)     & $26.59$ & \bad{$-4.20$}  & \good{$27.45$} & \good{$-0.24$}  & $26.34$        & \good{$-0.18$}  \\
\bottomrule
\end{tabular}
\end{table}

%Closing the gap does not cost accuracy in normal use, with matched supports: there LUC performs better than the paired model.
\textbf{LUC improves matched-support performance.} Importantly, the reduction in pairing dependence is not achieved by trading off performance under correctly paired supports; LUC instead improves matched-support performance. Across all five trained task types, LUC maintains or improves matched-support performance relative to the paired model (\cref{tab:both}).
On the primary whole-tumor segmentation task, \LU{} increases the matched-support DSC from $0.733$ to $0.857$ (subject-level Wilcoxon $p{=}2.4{\times}10^{-5}$).
%, on all five trained task types it is at or above that of the paired model (\cref{tab:both}), on
The improvement also extends across populations, with DSC on ADNI increasing from $0.466$ to $0.737$, and UniverSeg keeps its matched-support DSC.

\begin{figure}[t]\centering
\includegraphics[width=\textwidth]{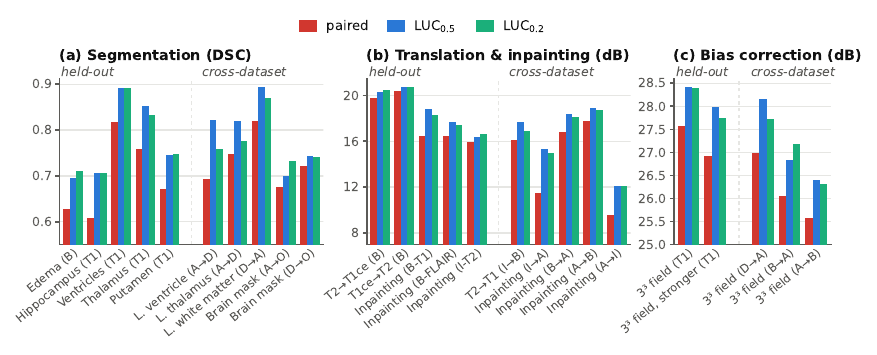}
\caption{\textbf{Performance on tasks outside training.}
Matched-support score on held-out tasks and on cross-dataset episodes, in which the supports and the query come from different datasets.
Held-out segmentation tasks use new combinations of anatomical structures that are segmented individually during training, while  the bias-field-correction task in (c) uses a coarser $3^3$ bias-field grid than in training.
Both LUC models score significantly above the paired model on every task shown (subject-level Wilcoxon, $p{<}0.05$); \cref{tab:heldout} in \cref{app:heldout-zs} lists all such tasks.
Datasets in parentheses: B: BraTS, A: ABCD, D: ADNI, O: OASIS-3, I: IXI; T1: the three T1 cohorts.}
\label{fig:heldout}
\end{figure}

The gain also holds outside the training distribution.
Held-out tasks change the segmentation target, the translation direction or the inpainting setting to ones that never occur in training, and cross-dataset episodes take the supports and the query from different datasets, as when a support set was collected on another scanner or population, whereas every training episode draws both from one dataset.
On $46$ of $53$ such tasks, both LUC models have a significantly higher matched-support score than the paired model, with near-zero gaps.
\Cref{fig:heldout} shows $25$ of these tasks, and \cref{tab:heldout} in \cref{app:heldout-zs} lists all $53$.
Notably, for a held-out bias-correction task with a coarser bias field  than those seen during training, the paired model barely corrects the input while both LUC models do (\cref{fig:heldout}c).

\begin{figure}[t]\centering
\includegraphics[width=\textwidth]{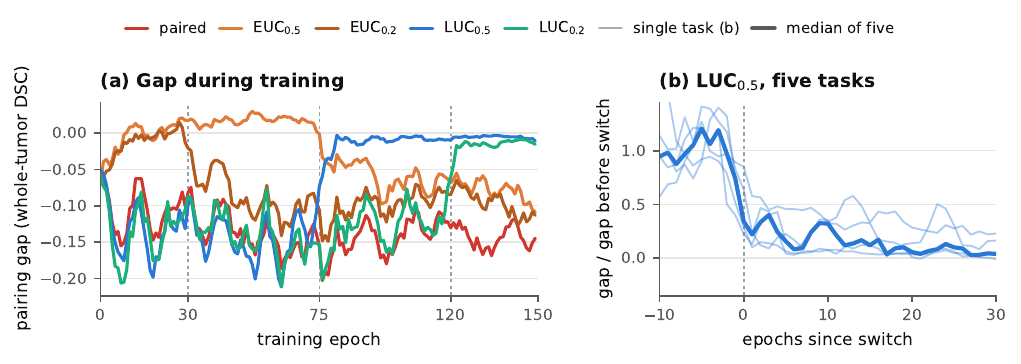}
\caption{\textbf{Pairing gap during training, and how fast it closes after the switch.}
Validation gap logged at every epoch.
(a) Whole-tumor gap, five-epoch moving average; dotted lines mark the schedule switches.
(b) Gap on each of five tasks (BraTS whole tumor, ABCD, ADNI, OASIS-3, T2$\to$FLAIR), divided by its mean over the ten epochs before the switch, for \LU{}; three-epoch moving average, thin lines for individual tasks and the thick line for their median.
\Cref{fig:gap-dynamics-full} in \cref{app:order} shows that \LUd{} behaves the same way.}
\label{fig:gap-dynamics}
\end{figure}

\textbf{The timing of unpaired training matters.}
Unpairing in every epoch does not enjoy the same combination of low pairing dependence and high matched-support performance as LUC: the unpaired model reaches a matched-support DSC of only $0.611$, against $0.733$ for the paired model (\cref{fig:both}).
%, although on a single task random unpairing throughout training closes the gap at the same matched-support DSC (\cref{app:shortcut} reports this single-task study).
On the other hand, EUC (trained on unpaired supports first and matched supports last) cannot fully close the gap; e.g., \EU{} and \EUd{} keep whole-tumor gaps of $-0.091$ and $-0.130$ (\cref{fig:both}).
The gap logged at every training epoch (\cref{fig:gap-dynamics}) shows that \LU{} reduces the pairing gap within a few epochs after switching to unpaired supports on all five tasks tracked during training, whereas both EUC models regain their gap once training returns to matched supports.
A short final phase of unpairing is therefore enough, and it adds no parameters, training steps, labels or inference cost.

\begin{figure}[t]\centering
\includegraphics[width=\textwidth]{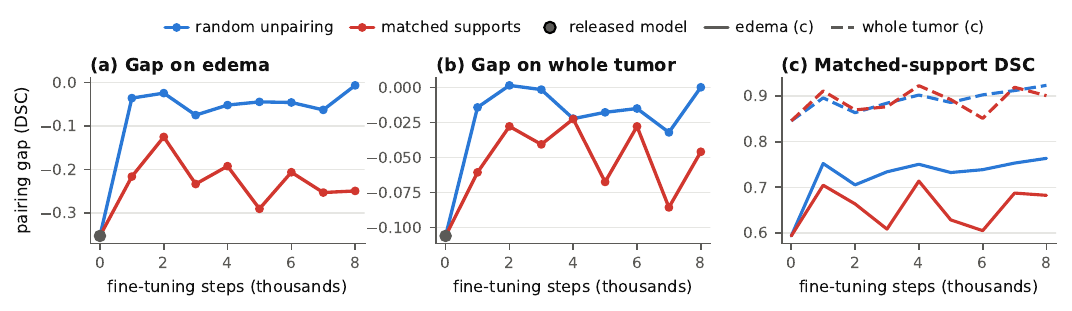}
\caption{\textbf{Fine-tuning the released Neuroverse3D of \cref{tab:generic-shortcut}, with random unpairing or with matched supports.}
Pairing gap on edema (a) and on whole tumor (b), and matched-support DSC (c), against fine-tuning steps.
Step 0 is the released model; every evaluation uses the same queries, supports and derangements as \cref{tab:generic-shortcut}.}
\label{fig:finetune-released}
\end{figure}

Therefore, for a released model the matched phase has already been run by its authors, so a short phase of random unpairing on top of it completes an LUC schedule.
We test this by fine-tuning the released Neuroverse3D on our data mix with random unpairing. As a control, we separately fine-tune the model with matched supports for the same number of steps (\cref{fig:finetune-released}).
After $1000$ steps of fine-tuning, a small fraction of one training run of our recipe, the gap for edema segmentation shrinks from $-0.353$ to $-0.036$ and stays near zero, while the control keeps a large gap; matched-support DSC rises in both arms (\cref{app:results} gives the values over the whole fine-tuning run).

\subsection{LUC reduces failure modes}\label{sec:fix}
\LU{} reduces each failure mode of \cref{sec:failure}, and \cref{fig:failure-modes} compares \LU{} and the paired model on the same cases and supports.
With real support sets, \LU{} adds fewer regions that the query does not have ($0.62$ per case against $2.40$).
Removing these extra regions hardly impacts the DSC of \LU{} ($0.007$, against $0.041$ for the paired model).
Moreover, with \LU{} the predicted tumor size hardly follows the tumor size in the support labels.
%About a third of its DSC advantage on these cases comes from not adding extra regions: deleting them would raise the DSC of the paired model by $0.041$ and its own by only $0.007$.
Under the controlled perturbations, shifting the support labels barely moves its prediction ($0.25$ voxels for \LU{} against $0.67$ for the paired model), and the shift also costs \LU{} only $0.001$ DSC, against $0.035$ for the paired model.
Lastly, an injected mask raises the predicted fraction of the mask by only $0.095$, against $0.846$ for the paired model (\cref{app:mechanism} gives the other models).
The timing of the unpairing matters here too: \EU{}, which ends on matched supports, still predicts most of the injected mask ($0.817$; \cref{fig:recency} in \cref{app:order}).
Support-specific properties thus carry over far less to the query predictions of \LU{}.

\section{Discussion}\label{sec:discussion}
\paragraph{The derangement as a test for other models.}
Any in-context learner that conditions on a set of examples~\citep{brown2020language, wang2023seggpt} draws on two sources of information: the correspondence within each example, and the examples as a set.
Deranging the support labels disrupts the first and holds the second fixed, and in the paired model it costs more DSC than replacing the labels with other patients' labels, which also changes which labels the set contains (\cref{app:shortcut}).
The derangement needs no new labels or held-out split and runs on a released model as it is.
Pairing dependence appears in all four released models, across 2D and 3D, four ways of fusing the supports and a general-purpose model, and its strength differs between models and tasks (\cref{tab:generic-shortcut}), so the derangement can be added to the evaluation of any visual ICL model.

\paragraph{Where the pairing dependence comes from, and how to regulate it.}
Under matched supports every label is correct for its own image, so relying on the pairing works throughout paired training.
The dependence follows the current phase of training: the gap closes within a few epochs of a switch to unpaired supports and comes back at a similar rate once training returns to matched supports (\cref{fig:gap-dynamics}).
As in curriculum learning~\citep{bengio2009curriculum}, the order of training matters, and for the gap it matters more than the amount of unpairing: \LUd{}, which unpairs only in the final fifth, has a gap near zero, while \EU{}, which unpairs for the whole first half, keeps a gap (\cref{fig:both}).
A practitioner therefore sets the dependence by choosing when unpairing starts, and a released model needs only a short final phase of unpairing (\cref{fig:finetune-released}).

\paragraph{Support-label errors in medical imaging.}
Support labels in medical in-context learning are few and drawn by hand~\citep{karimi2020deep}, and a label can end up out of register with its scan.
In the paired model, mis-registered support labels move the prediction and lower its accuracy (\cref{sec:failure}).
LUC reduces reliance on individual support labels while maintaining or improving matched-support performance: the same shift barely moves its prediction and costs almost no accuracy (\cref{sec:fix}).

\section{Conclusion}\label{sec:conclusion}
Visual in-context learners infer a task from support pairs, and we asked how strongly their predictions depend on the individual image--label pairings.
Deranging the support labels at test time leaves the label multiset unchanged and moves only the pairing, and the resulting pairing gap shows that this dependence varies widely across four released models.
The late unpairing curriculum, random unpairing scheduled late in training, substantially reduces the dependence and the failure modes of paired training while maintaining or improving matched-support performance across all evaluated task types, and a short unpairing phase shrinks the gap of a released model.
The reduction in the gap holds across three populations, on held-out tasks and on cross-dataset episodes.
With the same number of unpairing epochs, the schedule that ends on unpaired supports has the gap closer to zero, so the phase that ends training sets how much the prediction depends on the pairing.
The derangement diagnoses pairing dependence, and the schedule $p(e)$ controls it.

% \section*{Ethics statement}
% This work uses public, de-identified brain MRI datasets and collects no new data from human subjects.
% The failure modes we report, a prediction that follows the support labels rather than the query, matter for clinical use, and the training schedule we describe reduces them without extra annotation.

% \section*{Reproducibility statement}
% Every Neuroverse3D model we trained uses one recipe (\cref{sec:method}) in which only the unpairing schedule $p(e)$ varies, and the model, learning-rate schedule, data, task mix, optimizer, gradient clip, batch size, steps per epoch, seed, epoch budget and hardware were held fixed across runs.
% \Cref{app:impl} gives the training protocol, the unpairing schedules, the evaluation protocol for the released models, and the definition of every probe, together with the subject-level evaluation and its bootstrap confidence intervals.
% Code to reproduce our results is available at \url{https://anonymous.4open.science/r/Unpair-Med-vICL}.

% \section*{AI use statement}
% Generative AI tools helped the authors draft and polish the manuscript prose and write the figure-plotting and analysis-formatting code, under close author direction.
% The authors designed the experiments and produced the results.
% Every experiment, number, table, and figure was produced, checked, and verified by the authors, who take full responsibility for the content.

%% file: sections/appendix.tex
% ------------------------------------------------------------
\section{Implementation and evaluation details}\label{app:impl}
% ------------------------------------------------------------
This appendix describes the controlled training family, the data and preprocessing, the support-set operators, the evaluation of the released models, and the statistical procedures used in \cref{sec:method}.

\subsection*{Architecture and training recipe}
The backbone is Neuroverse3D~\citep{hu2024neuroverse3d}, a five-stage 3D U-Net~\citep{ronneberger2015u, cciccek20163d} with channels $[32,64,128,256,512]$ and $70.9$M parameters.
It receives a query image and $K$ support (image, label) pairs and predicts a dense output for the query.
The six models of the controlled family use Adam with learning rate $10^{-4}$ and no weight decay, loss $50\times\mathcal{L}_{\text{smooth-}L_3\text{-}L_1}$, gradient clipping to norm $0.25$, batch size $1$, fp32 precision, $2000$ steps per epoch, and $150$ epochs with seed $42$.
Flip, Sobel filtering, and the GIN intensity augmentation of \citet{ouyang2022causality} are each applied at a rate of $0.05$.
Training uses three supports; evaluation uses $K{=}4$ unless a sweep is stated.
All settings in \cref{tab:impl_recipe}, including the learning-rate schedule and hardware, are fixed across the family; only $p(e)$ varies.

\begin{table}[ht]\centering\small
\caption{\textbf{Training recipe for the controlled Neuroverse3D family.}
All six models share these settings and differ only in the unpairing schedule $p(e)$.}
\label{tab:impl_recipe}
\setlength{\tabcolsep}{4pt}
\resizebox{\textwidth}{!}{%
\begin{tabular}{ll}
\toprule
Ingredient & Value (fixed across the family) \\
\midrule
Architecture & Neuroverse3D, channels $[32,64,128,256,512]$, $\sim$71M params \\
Data mix & BraTS\,2021 $+$ three T1 datasets (ABCD, ADNI, OASIS-3) \\
Tasks & segmentation, translation, anatomical seg., inpainting, bias correction \\
Optimizer & Adam, lr $10^{-4}$, no weight decay \\
Learning-rate schedule & held fixed across the family \\
Loss & $50\times\mathcal{L}_{\text{smooth-}L_3\text{-}L_1}$ \\
Gradient clip / batch / steps-per-epoch & 0.25 (norm) / 1 / 2000 \\
Precision / support-set size & fp32 / 3 (train), $K{=}4$ (eval) \\
Augmentation & flip / Sobel / GIN-contrast, each at rate $0.05$ \\
Seed / epoch budget / hardware & 42 / 150 / uniform GPU \\
\textbf{Varies} & \textbf{unpairing schedule $p(e)$ only} \\
\bottomrule
\end{tabular}}
\end{table}

\subsection*{Random unpairing}
Each epoch is either matched ($p(e){=}0$), with every support keeping its own label, or unpairing ($p(e){=}1$).
In an unpairing epoch every support keeps its image and receives a copy of the label of a uniformly chosen other support.
All replacement labels come from the same episode, so one label can appear twice while another is dropped: random unpairing does not preserve the label multiset, whereas the test-time derangement below does.
The query and its target are never changed, so the loss is always computed against the correct answer.
A schedule is summarized by $x$, the fraction of the $E{=}150$ training epochs with $p(e){=}1$.

\subsection*{The model family}
The family contains six models (\cref{tab:recipe-sched}): the paired model ($p(e){=}0$ throughout), the unpaired model ($p(e){=}1$ throughout), and two time-reversed pairs.
\EU{} and \LU{} switch at epoch $75$; \EUd{} and \LUd{} place $30$ unpairing epochs at the beginning or at the end of training.
Within each pair the number of unpairing epochs is the same, so the reversal changes only when the unpairing happens.

\begin{table}[ht]\centering\small
\caption{\textbf{The six models.}
All share the recipe of \cref{tab:impl_recipe} and differ only in $p(e)$; $x$ is the fraction of epochs with unpairing, and ``mirror'' names the time-reversed partner.}
\label{tab:recipe-sched}
\setlength{\tabcolsep}{4pt}
\begin{tabular}{llcl}
\toprule
model & $p(e)$ & $x$ & mirror \\
\midrule
paired & $0$ for all $e$ & 0 & none \\
unpaired & $1$ for all $e$ & 1 & none \\
\EU{} & $1$ for $e{<}75$, else $0$ & 0.5 & \LU{} \\
\LU{} & $0$ for $e{<}75$, else $1$ & 0.5 & \EU{} \\
\EUd{} & $1$ for $e{<}30$, else $0$ & 0.2 & \LUd{} \\
\LUd{} & $0$ for $e{<}120$, else $1$ & 0.2 & \EUd{} \\
\bottomrule
\end{tabular}
\end{table}

\subsection*{Data and preprocessing}
Training uses BraTS\,2021~\citep{menze2014multimodal, bakas2017advancing, baid2021rsna}, with co-registered T1, T1ce, T2 and FLAIR images and nested ET/TC/WT labels, together with three T1 datasets, ABCD~\citep{casey2018adolescent}, ADNI~\citep{jack2008adni} and OASIS-3~\citep{lamontagne2019oasis}, which carry SynthSeg anatomical labels~\citep{billot2023synthseg}.
Preprocessing clips intensities to the $[0.5\%,99.5\%]$ percentiles, applies z-score normalization, then crops, pads and resizes each volume to $128^3$; images receive a final min--max normalization.
Labels use the crop coordinates of their images, so every support image and its label share one spatial frame, and labels are never intensity-normalized.
We use a $90/10$ train/validation split per dataset, and all reported numbers are computed on held-out validation subjects.
Whole-tumor labels follow the BraTS convention~\citep{menze2014multimodal}.

\subsection*{Released models and fine-tuning}
\Cref{tab:generic-shortcut} evaluates each released model as distributed, without fine-tuning.
Queries are FLAIR images of BraTS validation subjects, and supports are drawn from the training split.
Neuroverse3D and Medverse take 3D volumes with $K{=}4$ supports, on $20$ and $12$ validation subjects.
UniverSeg and SegGPT take one tumor-containing axial FLAIR slice per subject at $128{\times}128$, which SegGPT resizes to $448{\times}448$.
UniverSeg uses its default of $16$ supports on $125$ validation slices ($123$ with edema), and SegGPT uses $K{=}4$ on $24$ slices.
Each query gets one support draw and one derangement, shared by its matched and shuffled runs.
The released Neuroverse3D is also fine-tuned on our data mix at learning rate $10^{-5}$, with random unpairing at $p{=}1$ or with matched supports.
The fine-tuned model of \cref{fig:finetune-released} is scored every $1000$ steps on the same queries, supports and derangements as the released model.
This fine-tuning is separate from the controlled family.

\subsection*{The derangement and the support-set size}
The test-time operator draws a permutation $\sigma$ of the $K$ labels by rejection sampling until $\sigma(i)\neq i$ for all $i$, so no label stays on its own image.
It leaves the query, the support images and the label multiset unchanged.
A new support set is drawn for each query and held fixed across its matched and shuffled runs.
The pairing gap is
\[
  \Delta=M_{\text{shuffled}}-M_{\text{matched}},
\]
where $M$ is DSC or PSNR; a negative gap means the matched supports score higher, a positive gap that the shuffled supports do.
A derangement exists for $K\ge2$.
At $K{=}1$ the only permutation is the identity, so matched and shuffled supports are the same input and the gap is exactly zero, a built-in bookkeeping check.
The default is $K{=}4$.

\subsection*{The injected-mask test}
We add a fixed binary sphere to every support label: radius $9$ voxels, $3071$ voxels in total, centered at grid location $(44,44,64)$.
We discard any query whose true tumor overlaps the sphere by more than $2\%$, so the sphere is never a correct answer.
The injected-mask response is the change in the fraction of sphere voxels predicted as foreground,
\[
  R=P(\mathrm{pred}>0.5\mid\mathrm{sphere})_{\mathrm{after}}-P(\mathrm{pred}>0.5\mid\mathrm{sphere})_{\mathrm{before}},
\]
averaged over $n{=}25$ subjects.
A model that reproduces the support labels predicts the sphere and has a large response; a model that reads the answer from the query has a response near zero.

\subsection*{Subject-level evaluation}
Metrics are aggregated to one value per subject before any test, so no subject contributes more than once.
Unless stated otherwise, confidence intervals use a subject-level bootstrap with $5{,}000$ resamples; the support-volume regression of \cref{sec:failure} uses $10{,}000$ resamples of the queries.
Within-subject comparisons (matched against shuffled supports, or one model against another) use the Wilcoxon signed-rank test, with Benjamini--Hochberg FDR control within each task family.
The multi-dataset evaluation uses $n{=}30$ subjects for segmentation, $n{=}20$ for translation and $n{=}12$ for anatomical segmentation; the failure-mode analysis uses $n{=}60$, and each probe states its own sample size.
The pairing gap is a within-model, within-subject contrast, so each model is compared with itself.

\subsection*{Released code}
Code to reproduce our results is available in our
\href{https://github.com/JornyWan/pairing-shortcuts-3D-med-visual-ICL}{project repository}.
% ------------------------------------------------------------
\section{Additional diagnostics of pairing dependence}\label{app:shortcut}
% ------------------------------------------------------------
This appendix reports label-perturbation tests that place the gap in the assignment of labels to images rather than in the label content.

\subsection*{Pairing-specific controls}
We evaluate the multi-task models under matched, cross-patient, zeroed, Gaussian-noise and shuffled support labels ($K{=}4$, $n{=}30$; \cref{tab:disc-controls}).

The contrast that matters is shuffled against zeroed.
With shuffled supports, each support of the paired model ($p{=}0$) carries another support's label from the same support set.
Every one of those labels is real and correct for some image in the set.
Moving the labels this way costs more than zeroing them: $0.549$ shuffled $<0.655$ zeroed $<0.661$ cross-patient $<0.733$ matched.
A model sensitive to any change in the labels would lose at least as much when the labels are zeroed as when they are moved among real images.
The measured order is the reverse, which places the loss in the assignment.
Gaussian-noise labels, which destroy the label content, score lower still ($0.461$).
We therefore read pairing specificity from the ordering against the zeroed and the cross-patient condition: the paired model depends on which label goes with which image.

The LUC models move little between matched and shuffled supports, $0.857$ against $0.848$ for \LU{} and $0.878$ against $0.875$ for \LUd{}, and they hold the highest matched-support DSC in the table.
Both EUC models, which end on matched supports, lose DSC under shuffling like the paired model ($0.731$ to $0.640$ and $0.731$ to $0.601$).
The unpaired model scores $0.611$ with matched supports and $0.719$ with shuffled supports; it was trained only on unpaired supports, so matched ones are new to it.
Across the models in this table, pairing dependence follows the training distribution.

\begin{table}[h]\centering\small
\caption{\textbf{Label-perturbation tests on BraTS segmentation} (subject-level mean DSC, $n{=}30$).
Shuffling changes only the assignment; cross-patient, zeroed and noise labels also change the label content.
For the paired model, shuffled supports score below the zeroed and the cross-patient condition; the LUC models move little between matched and shuffled supports and hold the highest matched-support DSC.}
\label{tab:disc-controls}
\begin{tabular}{lccccc}
\toprule
model & matched & cross-patient & zero & noise & shuffled \\
\midrule
paired         & 0.733 & 0.661 & 0.655 & 0.461 & \textbf{0.549} \\
unpaired       & 0.611 & 0.713 & 0.714 & 0.573 & 0.719 \\
\LU{}    & \textbf{0.857} & 0.832 & 0.826 & 0.710 & \textbf{0.848} \\
\LUd{}      & \textbf{0.878} & 0.878 & 0.879 & 0.808 & \textbf{0.875} \\
\EU{} & 0.731 & 0.669 & 0.631 & 0.541 & 0.640 \\
\EUd{} & 0.731 & 0.646 & 0.650 & 0.534 & 0.601 \\
\bottomrule
\end{tabular}
\end{table}

% ------------------------------------------------------------
\section{The failure modes in detail}\label{app:failure}
% ------------------------------------------------------------
The measurements of \cref{sec:failure,sec:fix} use the paired model and \LU{} on the same $60$ BraTS validation subjects, with the same real support images and support labels ($K{=}4$).

\paragraph{Extra regions.}
We count predicted connected components of at least $50$ voxels that overlap the query ground truth by less than $10\%$.
The paired model produces $2.40$ per case against $0.62$ for \LU{} (subject-level Wilcoxon $p{=}1.3{\times}10^{-4}$), and $62\%$ of its cases carry at least one against $30\%$.
Deleting those regions raises matched-support DSC by $0.041$ for the paired model and by $0.007$ for \LU{}, so about a third of the $0.114$ DSC difference between the two models comes from this one failure mode.
We also recorded where those components fall.
Of their voxels, $0.35$ lie inside the union of the four support labels for the paired model, against $0.27$ for the union of four labels drawn from patients outside the support set.
The extra regions therefore fall more often where the support labels carry tumor.

\paragraph{Predicted size.}
We regressed predicted volume on true volume and mean support volume together, so the support term measures the effect of the support volume after accounting for the true volume.
Its coefficient is $0.460$ (95\% CI $[0.026,0.990]$) for the paired model and $0.122$ ($[-0.094,0.346]$) for \LU{}, a difference of $0.338$ ($[0.037,0.720]$) from $10{,}000$ bootstrap resamples of the queries, so larger support labels go with a larger prediction about four times as strongly for the paired model.

\paragraph{Mis-registered supports.}
Shifting every support label by $8$ voxels along each of two axes moves the paired model's predicted centroid $0.67$ voxels along the shift direction, against $0.25$ voxels for \LU{} ($p{=}0.022$).
\Cref{fig:safety} in \cref{app:results} gives the DSC this shift costs, together with replaced and eroded support labels.

\subsection*{A real-support example and the injected-mask test}
\begin{figure}[t]\centering
\includegraphics[width=\textwidth]{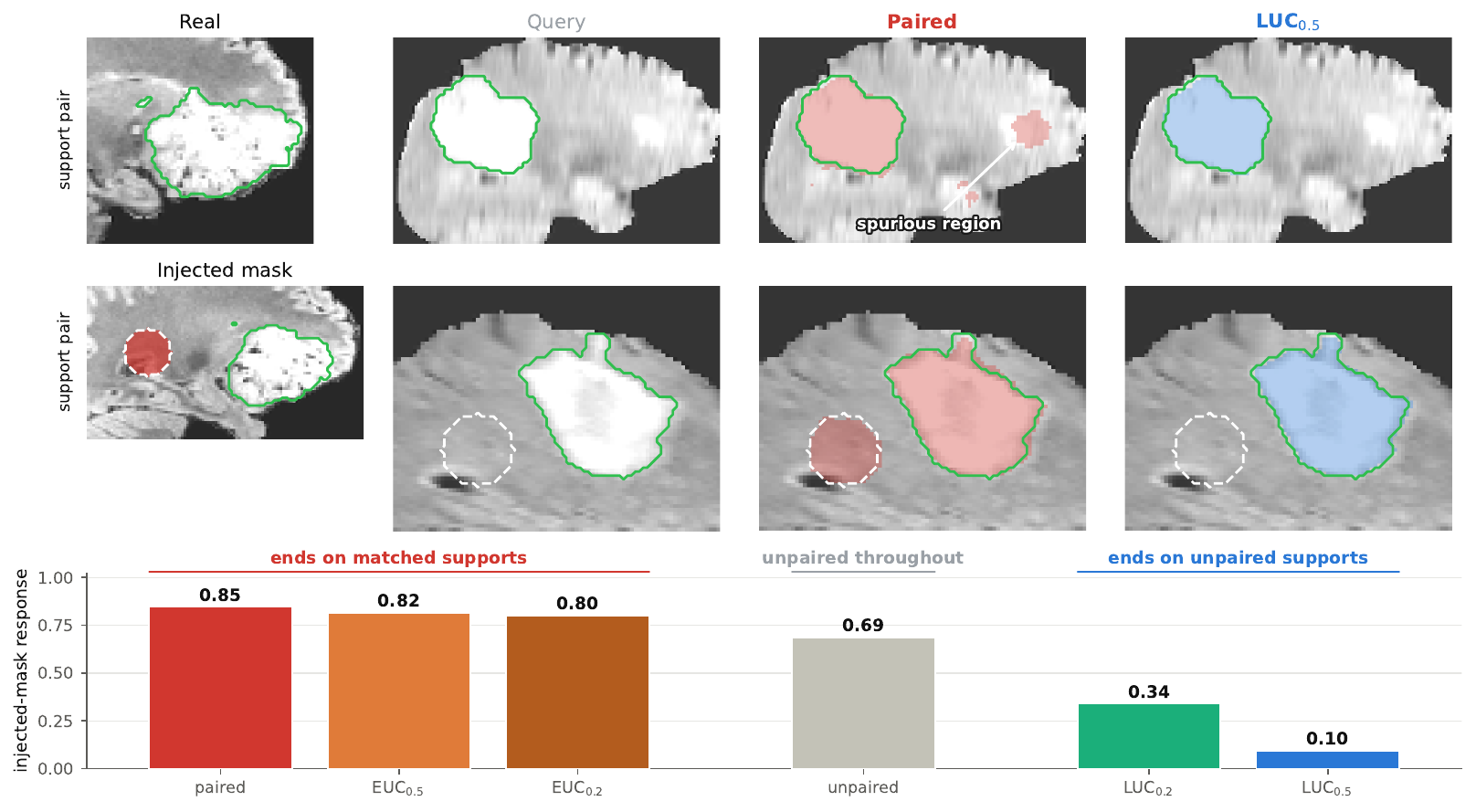}
\caption{\textbf{Predictions with real supports and with a mask painted into the supports.}
Each row shows a support pair, the query, and the paired and \LU{} predictions.
Top: real supports; the query has a single tumor (green).
Middle: the injected-mask test, with a fixed synthetic label (red, white dashed ring) painted into every support where the query has no tumor.
Bottom: injected-mask response across the models, grouped by schedule.}
\label{fig:leakage}
\end{figure}

The effect appears with real supports and no manipulation (\cref{fig:leakage}, top).
On a query whose ground truth is a single tumor, the paired model predicts two disconnected regions.
It places one of them where the supports carry tumor and the query does not, while \LU{} predicts only the real region.
This over-segmentation is systematic: across single-tumor queries the paired model adds a spurious disconnected region in $71\%$ of cases, against $38\%$ for \LU{}.
We then paint a fixed synthetic mask into all $K$ support labels at a location outside the query's tumor, so the only change to the input is that mask (\cref{fig:leakage}, middle; definition in \cref{app:impl}).
The paired model reproduces the mask in its own prediction (injected-mask response $+0.846$), while the response of \LU{} is $+0.095$.
\Cref{app:mechanism} gives the response across the family.

% ------------------------------------------------------------
\section{Complete results for LUC}\label{app:results}
% ------------------------------------------------------------
\paragraph{One evaluation pipeline.}
The task-level tables of this section (\cref{tab:app-id,tab:app-pop}) come from one evaluation that scores all six models with a single pipeline.
Matched-support scores and pairing gaps are therefore comparable across every row, and they are the numbers behind \cref{tab:both}.
Throughout, ``matched'' is the matched-support score and ``gap'' is the pairing gap, the shuffled-support minus the matched-support score, so a more negative gap means more pairing dependence.

\subsection*{In-distribution results across the five trained task types}
\Cref{tab:app-id} gives the matched-support score and the gap for every model and trained task type.
\LU{} brings the gap near zero on all five task types with matched-support scores at or above the paired model ($0.857$, $22.557$, $0.806$, $17.756$ and $27.447$ against $0.733$, $22.406$, $0.704$, $17.517$ and $26.588$).
This is the result of \cref{sec:both}, read task by task.
\LUd{} also brings the gap near zero on all five task types and reaches $0.878$ on segmentation.
The unpaired model loses at most $0.011$ under the derangement on every task type, with matched-support segmentation DSC $0.611$.
\LU{} keeps matched supports through the first half of training and reaches $0.857$.

\begin{table}[h]\centering\scriptsize
\caption{\textbf{In-distribution results (matched\,/\,gap) for the six models trained under the recipe.}
Matched $=$ the score with matched supports, in the metric named in the column header; gap $=$ shuffled $-$ matched, so a more negative gap means more pairing dependence.
\good{Green}: the LUC models, where the gap is near zero and the matched-support score is at or above the paired model.
\bad{Red}: the paired model's gap.
All rows share the one recipe and differ only in $p(e)$.}
\label{tab:app-id}
\resizebox{\textwidth}{!}{%
\begin{tabular}{lccccc}
\toprule
model & Seg DSC & Transl PSNR & AnatSeg DSC & Inp PSNR & Bias PSNR \\
\midrule
paired   & 0.733 / \bad{$-0.184$} & 22.406 / \bad{$-1.216$} & 0.704 / \bad{$-0.035$} & 17.517 / \bad{$-2.267$} & 26.588 / \bad{$-4.201$} \\
unpaired & 0.611 / $+0.109$ & 22.146 / $+0.042$ & 0.685 / $+0.005$ & 17.327 / $-0.011$ & 25.958 / $+0.159$ \\
\EU{} & 0.731 / $-0.091$ & 22.389 / $-1.099$ & 0.708 / $-0.024$ & 17.339 / $-0.880$ & 25.661 / $-3.234$ \\
\LU{} & \good{0.857} / \good{$-0.008$} & \good{22.557} / \good{$-0.027$} & \good{0.806} / \good{$-0.001$} & \good{17.756} / \good{$+0.052$} & \good{27.447} / \good{$-0.242$} \\
\EUd{} & 0.731 / $-0.130$ & 22.486 / $-1.340$ & 0.694 / $-0.025$ & 17.651 / $-1.872$ & 26.516 / $-3.759$ \\
\LUd{} & \good{0.878} / \good{$-0.003$} & \good{22.698} / \good{$-0.081$} & \good{0.791} / \good{$-0.007$} & \good{17.878} / \good{$-0.021$} & 26.338 / \good{$-0.182$} \\
\bottomrule
\end{tabular}}

\end{table}

\paragraph{Inpainting and bias correction.}
The paired model has its largest gaps on the two restoration tasks, $-2.267$\,dB for inpainting and $-4.201$\,dB for bias correction.
\LU{} has $+0.052$ and $-0.242$\,dB, and \LUd{} has $-0.021$ and $-0.182$\,dB.
As on segmentation and translation, the schedules that end on unpaired supports have the gaps closer to zero.

\subsection*{Cross-population results}
\Cref{tab:app-pop} splits the anatomical-segmentation result of \cref{tab:both} into its three cohorts.
On all three cohorts \LU{} has a gap within $0.004$ of zero and \LUd{} within $0.016$, and both have matched-support DSC above the paired model.
On the two cohorts where the paired model has a gap (ABCD and ADNI), \EU{} and \EUd{} have more negative gaps than \LU{} and \LUd{}, the same order effect as in \cref{app:order}.
The matched-support gain of \LU{} over the paired model is largest on ADNI ($0.466$ to $0.737$), and on OASIS-3 all five models have a gap of zero to three decimals.
The pairing gap and its reduction under LUC therefore extend from BraTS whole tumor to anatomical segmentation in children (ABCD) and older adults (ADNI).

\begin{table}[h]\centering\small
\caption{\textbf{Pairing gap by cohort (anatomical segmentation, $K{=}4$).}
Matched $=$ matched-support DSC; gap $=$ shuffled-support minus matched-support DSC.
\good{Green}: matched-support DSC under LUC at or above the paired model.
\bad{Red}: the paired model's gap on ABCD.
On all three cohorts the LUC models have a gap within $0.016$ of zero and matched-support DSC above paired; \EU{} and \EUd{} (which end on matched supports) have more negative gaps than \LU{} and \LUd{} where the paired model has one.}
\label{tab:app-pop}
\setlength{\tabcolsep}{4pt}
\resizebox{\ifdim\width>\linewidth\linewidth\else\width\fi}{!}{%
\begin{tabular}{l cc cc cc cc cc}
\toprule
 & \multicolumn{2}{c}{\textbf{Paired} ($p{=}0$)} & \multicolumn{2}{c}{\LU{}} & \multicolumn{2}{c}{\LUd{}} & \multicolumn{2}{c}{\EU{}} & \multicolumn{2}{c}{\EUd{}} \\
\cmidrule(lr){2-3}\cmidrule(lr){4-5}\cmidrule(lr){6-7}\cmidrule(lr){8-9}\cmidrule(lr){10-11}
\textbf{Cohort} & matched & gap & matched & gap & matched & gap & matched & gap & matched & gap \\
\midrule
ABCD (children)   & $0.859$ & \bad{$-0.074$} & \good{$0.868$} & $+0.002$ & \good{$0.871$} & $-0.006$ & $0.848$ & $-0.049$ & $0.857$ & $-0.048$ \\
ADNI (elderly)    & $0.466$ & $-0.030$        & \good{$0.737$} & $-0.004$ & \good{$0.702$} & $-0.016$ & $0.504$ & $-0.024$ & $0.438$ & $-0.026$ \\
OASIS-3 (mixed)    & $0.785$ & $0.000$        & \good{$0.813$} & $0.000$ & \good{$0.800$} & $0.000$ & $0.774$ & $0.000$ & $0.788$ & $0.000$ \\
\bottomrule
\end{tabular}}
\end{table}

\subsection*{Qualitative multi-task comparison}
\begin{figure}[h]\centering
\includegraphics[width=\textwidth]{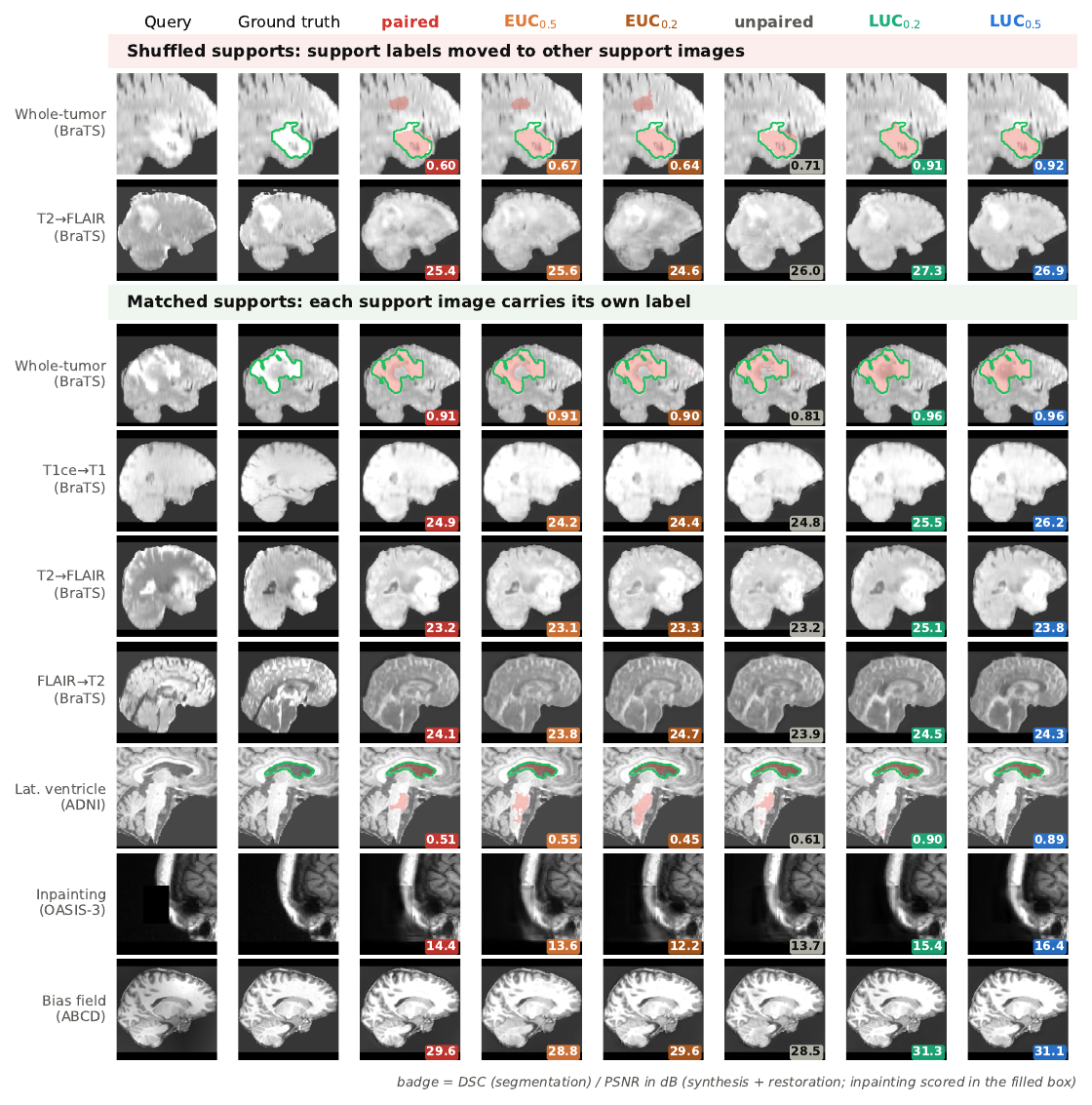}
\caption{\textbf{Qualitative comparison across tasks and models ($K{=}4$).}
Six models (paired $\to$ \EU{} $\to$ \EUd{} $\to$ unpaired $\to$ \LUd{} $\to$ \LU{}); green $=$ ground-truth contour, red $=$ predicted mask; the corner badge is DSC (segmentation) or PSNR in dB (synthesis/restoration; inpainting scored in the filled box only).
Rows are labeled task\,(dataset).
\textbf{Top, shuffled supports:} two tasks; the paired model degrades (partial or displaced mask, lost detail) while \LU{} and \LUd{} hold, \EU{} and \EUd{} degrade with the paired model, and the unpaired model sits in between.
\textbf{Bottom, matched supports:} the trained task suite across all four datasets.
Accuracy is comparable throughout: \LU{} matches the paired model on translation and restoration, is cleaner on the two segmentation tasks (whole-tumor over-segmentation; atrophic ADNI ventricle), and fills the inpainting box slightly better.
The bottom-row cases were selected to show the pattern; population numbers are in \Cref{tab:app-id,tab:app-pop}.}
\label{fig:gallery}
\end{figure}

\Cref{fig:gallery} shows single cases behind the numbers.
The top block uses shuffled supports on two tasks.
On BraTS whole tumor the paired model returns a partial, displaced mask (DSC ${\sim}0.6$), and on T2$\to$FLAIR synthesis it loses tissue detail, while \LU{} and \LUd{} track the target.
\EU{} and \EUd{} degrade along with the paired model, and the unpaired model sits between the two.
The bottom block uses matched supports and spans the trained task suite across all four datasets: whole-tumor segmentation, three modality-translation directions, anatomical segmentation (ADNI), inpainting (OASIS-3) and bias-field correction (ABCD).
Accuracy there is comparable across the models.
\LU{} matches the paired model on translation and restoration and fills the inpainting box slightly better.
On the two segmentation tasks it drops the spurious whole-tumor region the paired model adds, and it recovers the atrophic ADNI ventricle the paired model under-segments.
Tasks outside training are quantified in \cref{app:heldout-zs}.

\subsection*{A second backbone: UniverSeg}
We trained UniverSeg from scratch on BraTS multi-region segmentation, where the support pair specifies which of WT, TC, ET and edema to segment, and added LUC to its own recipe (\cref{tab:generic-fix}).
The gap shrinks from $-0.222$ to ${\approx}0$ with matched-support DSC $0.590$ and $0.602$, so LUC carries over to a 2D backbone.
Trained from scratch on a single task, UniverSeg shows no gap we could measure.

\begin{table}[t]\centering\small
\caption{\textbf{Pairing gap with and without LUC, in a 3D and a 2D backbone.}
Each backbone is trained with its own recipe; its two rows differ only in the unpairing schedule $p(e)$.
Neuroverse3D: BraTS whole tumor; UniverSeg: BraTS multi-region, mean over WT/TC/ET/edema.}
\label{tab:generic-fix}
\begin{tabular}{llcc}
\toprule
backbone & training & matched DSC & pairing gap (DSC) \\
\midrule
\multirow{2}{*}{Neuroverse3D~\citep{hu2024neuroverse3d} (3D)} & paired      & 0.733 & $-0.184$ \\
                                   & \quad$+$ \LU{} & \textbf{0.857} & $-0.008$ \\
\midrule
\multirow{2}{*}{UniverSeg~\citep{butoi2023universeg} (2D)}    & paired      & 0.590 & $-0.222$ \\
                                   & \quad$+$ \LU{} & 0.602 & $\approx\!0.000$ \\
\bottomrule
\end{tabular}
\end{table}

\subsection*{LUC on the released Neuroverse3D}
\Cref{fig:finetune-released} fine-tunes the released Neuroverse3D with random unpairing and, as a control, with matched supports, following the protocol in \cref{app:impl}.
The first $1000$ steps of unpairing, $0.3\%$ of the $300{,}000$ steps in one training run of our recipe, take the edema gap from $-0.353$ to $-0.036$ and the whole-tumor gap from $-0.106$ to $-0.014$.
The gaps stay near zero for the remaining $7000$ steps (means over the eight evaluations $-0.043$ on edema and $-0.013$ on whole tumor), while the control keeps an edema gap of $-0.221$ on average, so the drop comes from the unpairing rather than from further training on our data.
Matched-support DSC rises in both arms, and on edema it is higher with unpairing at every evaluation ($0.741$ against $0.662$ on average).

\subsection*{Replaced, shifted and eroded support labels}
\begin{figure}[t]\centering
\includegraphics[width=\textwidth]{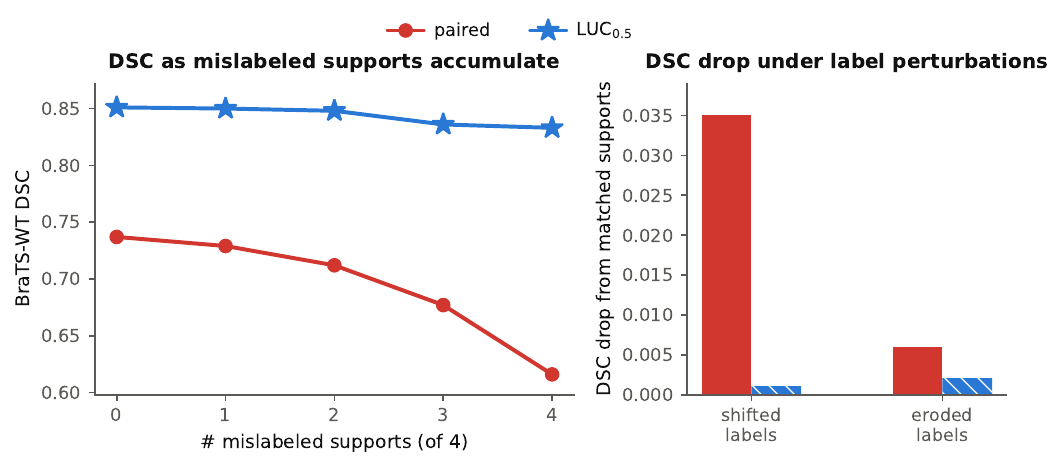}
\caption{\textbf{Score under replaced, mis-registered, and eroded support labels.}
Left: BraTS whole-tumor DSC as $0$ to $4$ of the $K{=}4$ support labels are replaced by another patient's label.
Right: DSC drop from the matched-support score under a support mis-registration and a partial erosion of the support masks.
Neuroverse3D, $n{=}60$.}
\label{fig:safety}
\end{figure}

The derangement is a controlled test, and we also measured support-set errors that occur in practice (\cref{fig:safety}).
As we replace $0$ to $4$ of the $K{=}4$ support labels with another patient's label, the paired model's BraTS whole-tumor DSC falls from $0.74$ to $0.62$, while \LU{} goes from $0.85$ to $0.83$.
A mis-registration of the support labels by $8$ voxels along each of two axes costs the paired model $0.035$ DSC and \LU{} $0.001$.

% ------------------------------------------------------------
\section{Schedule order}\label{app:order}
% ------------------------------------------------------------
\paragraph{Comparison design.}
The two time-reversed pairs of \cref{tab:recipe-sched} unpair for the same number of epochs and differ only in temporal order, and every other training setting is fixed across the family, so the two runs of a pair differ only in the timing of the unpairing.

\paragraph{The six models.}
\Cref{fig:both} gives matched-support DSC and the pairing gap of all six models on BraTS whole tumor at $K{=}4$.
With the same $x$, the gap follows whether training ends on matched or on unpaired supports, at $x{=}0.5$ and at $x{=}0.2$.
\EUd{}, whose $30$ unpairing epochs are followed by $120$ matched epochs, ends with a gap of $-0.130$: the gap comes back once training returns to matched supports.

\begin{figure}[t]\centering
\includegraphics[width=0.52\textwidth]{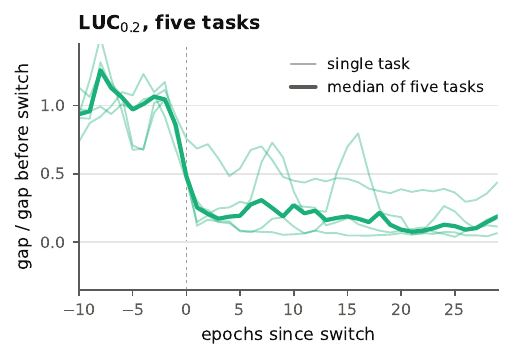}
\caption{\textbf{Pairing gap of \LUd{} after its switch at epoch $120$.}
The gap on each of five tasks, divided by its mean over the ten epochs before the switch; thin lines for tasks and the thick line for their median.}
\label{fig:gap-dynamics-full}
\end{figure}

\paragraph{Within-run trajectory.}
The per-epoch validation gap shows the timing directly (\cref{fig:gap-dynamics}; per-task trajectories of \LUd{} in \cref{fig:gap-dynamics-full}).
\EU{} has a gap of about $0$ through epoch $74$, and the gap opens right after its switch to matched supports at epoch $75$.
\LU{} carries the whole-tumor gap of the paired model before its switch ($-0.153$ over epochs $60$ to $74$); after the switch to unpaired supports at epoch $75$ the gap is within $0.02$ of zero in five epochs and averages $-0.007$ over epochs $90$ to $149$, closing about as fast as the gap of \EU{} opens.
\EUd{} has a gap of about $0$ during its unpaired phase, and the gap opens once training returns to matched supports.
The gap of \LUd{} shrinks from $-0.124$ to $-0.013$ within one epoch of its switch at epoch $120$.
On each of the five tasks tracked during training, the gap of \LU{} shrinks by $71$ to $97\%$ after the switch (\cref{fig:gap-dynamics}b), and \cref{fig:gap-dynamics-full} shows the same for \LUd{}.
Pairing dependence tracks the current training distribution: it follows the most recent phase and rises and falls at similar rates.

\begin{figure}[t]\centering
\includegraphics[width=\textwidth]{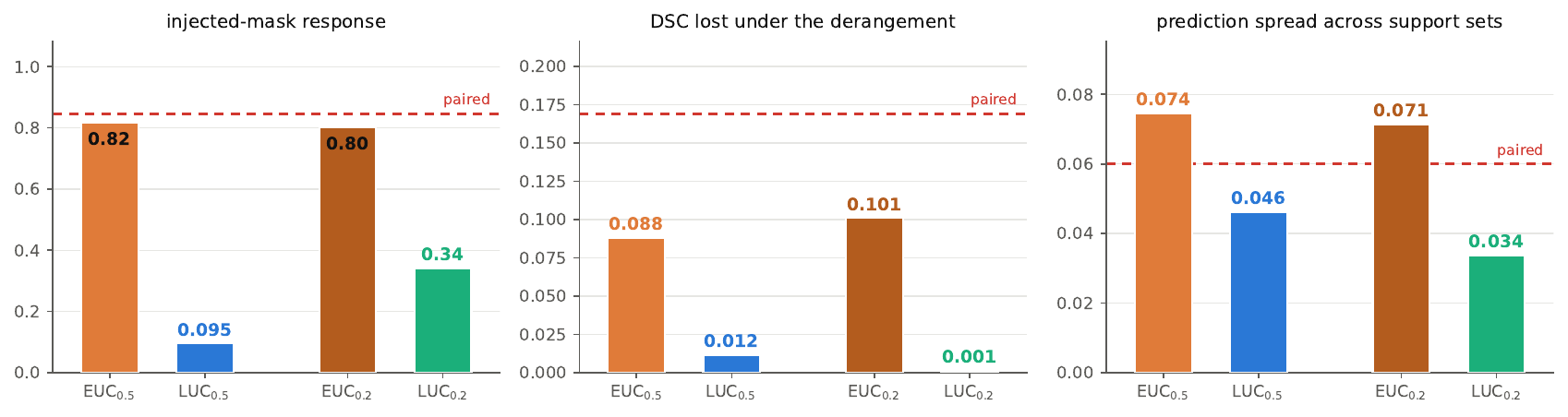}
\caption{\textbf{LUC versus EUC at $x{=}0.5$ and $x{=}0.2$.}
Injected-mask response, DSC lost under the derangement and support sensitivity; the two models of a pair unpair for the same number of epochs, the LUC model ends on unpaired supports and its EUC mirror on matched supports, and the dashed line is the paired model.}
\label{fig:recency}
\end{figure}

\paragraph{LUC and EUC on three probes.}
\Cref{fig:recency} compares the two mirror pairs on the injected-mask response, the DSC lost under the derangement (\cref{tab:spectrum}) and the support sensitivity, the per-voxel standard deviation of the prediction inside the true tumor across four independently drawn support sets.
For \LU{} against \EU{} these are $0.095$ against $0.817$, $0.012$ against $0.088$ and $0.046$ against $0.074$, and for \LUd{} against \EUd{} they are $0.340$ against $0.801$, $0.001$ against $0.101$ and $0.034$ against $0.071$.

% ------------------------------------------------------------
\section{Probes of how the models use the support set}\label{app:mechanism}
% ------------------------------------------------------------
The probes in this section measure how the predictions respond to perturbed support labels.
Both compare all six models, and each states its sample size.

\subsection*{Injected-mask response}
The injected-mask test adds the sphere of \cref{app:impl} to every support label and measures the change in the fraction of sphere voxels predicted as foreground ($n{=}25$; \cref{fig:leakage}).
The values order by what training ends on: paired $+0.846$, \EU{} and \EUd{} (end on matched supports) $+0.817$ and $+0.801$, \LUd{} $+0.340$, and \LU{} (ends on unpaired supports) $+0.095$.
The unpaired model responds with $+0.687$: unpairing throughout gives it a positive gap, yet it reproduces most of the injected mask, while \LU{} brings both close to zero.
The ordering puts the models that end on matched supports (paired and both EUC models) at the high end and the two LUC models at the low end.

\subsection*{Three support-label conditions}
\Cref{tab:spectrum} evaluates BraTS whole-tumor DSC ($n{=}40$) under three support-label conditions.
\texttt{correct} leaves the supports as they are.
\texttt{deranged} moves the labels to other support images, with the multiset unchanged.
\texttt{rolled16} shifts each label $16$ voxels relative to its image.
A model that depends on the assignment scores highest at \texttt{correct} and lower off it; a model that does not is flat across the three.
The paired model drops from $0.754$ to $0.585$ at \texttt{deranged}, and both EUC models behave the same way ($0.755\!\to\!0.667$ and $0.751\!\to\!0.650$).
Both LUC models are flat and high across all three conditions (${\ge}0.848$).
The unpaired model is lowest under correct labels and rises under derangement, because a matched support set is itself out-of-distribution for a model trained only on unpaired supports.

\begin{table}[h]\centering\small
\caption{\textbf{Three support-label conditions} (BraTS whole-tumor DSC, $n{=}40$).
The paired model and both EUC models lose DSC at \texttt{deranged}; \LU{} and \LUd{} are flat across all three support-label conditions.}
\label{tab:spectrum}
\begin{tabular}{lccc}
\toprule
model & correct & deranged & rolled16 \\
\midrule
paired     & 0.754 & 0.585 & 0.685 \\
unpaired   & 0.630 & 0.741 & 0.710 \\
\EU{}  & 0.755 & 0.667 & 0.686 \\
\EUd{} & 0.751 & 0.650 & 0.699 \\
\LU{}    & \textbf{0.860} & \textbf{0.848} & 0.852 \\
\LUd{}      & \textbf{0.880} & \textbf{0.879} & 0.880 \\
\bottomrule
\end{tabular}
\end{table}

The probes agree in direction.
The two LUC models, which end on unpaired supports after a matched phase, hold their DSC across the three support-label conditions and give the weakest injected-mask response.
The paired model and both EUC models, which end on matched supports, lose DSC when the labels move and reproduce the injected mask.
The two models of each mirror pair unpair for the same number of epochs and differ in when the unpairing is applied, so what separates them is what training ends on.
\LU{} combines a small pairing gap and a weak injected-mask response with matched-support DSC $0.857$ against $0.733$ for the paired model.

% ==== BEGIN tasks outside training (2026-09-24; generated table: Unpaired/scripts/heldout_submission_table.py) ====
\section{Tasks outside training}\label{app:heldout-zs}
We evaluate the paired model and both LUC models on tasks that never occur in training, with the protocol of \cref{app:results}: $K{=}4$ supports fixed per dataset from its training split, queries from its validation split (40 for BraTS and IXI\footnote{\url{https://brain-development.org/ixi-dataset/}}, 20 per T1 cohort), matched and shuffled supports, DSC for segmentation, and PSNR inside the target foreground or, for inpainting, inside the filled boxes.
One derangement is drawn per query and shared by all models.
Held-out tasks change the segmentation target, the translation direction, the inpainting setting or the bias-field grid, or apply inpainting to BraTS and IXI; the held-out segmentation targets span both hemispheres or several structures, while training used single structures.
Cross-dataset episodes take the supports from one dataset and the queries from another, which never happens in training.
\Cref{tab:heldout} lists $53$ tasks, grouped by task type: the tasks on which all three models perform the task, with DSC of at least $0.6$ or at least $1$\,dB above the unprocessed input, and at least one LUC model scores significantly above the paired model while neither scores significantly below it, together with three bias-correction tasks with a coarser $3^3$ control grid (on the T1 cohorts, BraTS$\to$ABCD and ABCD$\to$BraTS) on which the paired model stays within $0.2$\,dB of the input while both LUC models improve on it by $0.7$ to $1.3$\,dB.
On $46$ of the $53$ tasks both LUC models score significantly above the paired model (subject-level Wilcoxon, $p{<}0.05$); on the seven marked ones, one of them does and the other does not differ significantly from it.
On the $46$ tasks, \LU{} scores up to $0.128$ DSC and $3.8$\,dB above the paired model, and its pairing gap averages $-0.004$ DSC and $-0.13$\,dB, against $-0.054$ DSC and $-2.41$\,dB for the paired model.

\begin{table}[t]\centering\scriptsize
\caption{\textbf{Tasks outside training.} Matched-support score (DSC; PSNR in dB, inside the filled boxes for inpainting) and pairing gap; ``input'' is the PSNR of the unprocessed input, and an arrow in the data column marks a cross-dataset episode. Both LUC models score significantly above the paired model on every unmarked task (subject-level Wilcoxon, $p{<}0.05$); on tasks marked $^\circ$, one LUC model does and the other does not differ significantly from it. $^\dagger$Shown in \cref{fig:heldout}.}
\label{tab:heldout}
\setlength{\tabcolsep}{3.5pt}
\resizebox{\textwidth}{!}{%
\begin{tabular}{llc c cc cc cc}\toprule
task & data & $n$ & input & \multicolumn{2}{c}{paired ($p{=}0$)} & \multicolumn{2}{c}{\LU{}} & \multicolumn{2}{c}{\LUd{}} \\
\cmidrule(lr){5-6}\cmidrule(lr){7-8}\cmidrule(lr){9-10}
 & & & & matched & gap & matched & gap & matched & gap \\
\midrule
\multicolumn{10}{l}{\emph{Segmentation (DSC)}} \\
\quad Hippocampus$^\dagger$ & T1 cohorts & 60 & -- & 0.608 & $-$0.011 & 0.706 & $+$0.004 & 0.707 & $-$0.017 \\
\quad Caudate & T1 cohorts & 60 & -- & 0.632 & $-$0.014 & 0.720 & $-$0.004 & 0.716 & $-$0.014 \\
\quad Thalamus$^\dagger$ & T1 cohorts & 60 & -- & 0.758 & $-$0.001 & 0.853 & 0.000 & 0.832 & $-$0.002 \\
\quad Edema from FLAIR$^\dagger$ & BraTS & 40 & -- & 0.627 & $-$0.088 & 0.695 & $-$0.003 & 0.711 & $-$0.004 \\
\quad Putamen$^\dagger$ & T1 cohorts & 60 & -- & 0.672 & $-$0.019 & 0.746 & $-$0.007 & 0.748 & $-$0.013 \\
\quad Lateral ventricles$^\dagger$ & T1 cohorts & 60 & -- & 0.817 & $-$0.034 & 0.893 & $-$0.003 & 0.892 & $-$0.004 \\
\quad Ventricular system & T1 cohorts & 60 & -- & 0.792 & $-$0.042 & 0.856 & $-$0.005 & 0.854 & $-$0.009 \\
\quad Cerebral white matter & T1 cohorts & 60 & -- & 0.849 & $-$0.115 & 0.890 & 0.000 & 0.889 & $-$0.001 \\
\quad Cerebellum and brainstem & T1 cohorts & 60 & -- & 0.888 & $-$0.008 & 0.904 & 0.000 & 0.913 & $-$0.007 \\
\quad Cerebellum & T1 cohorts & 60 & -- & 0.872 & $-$0.007 & 0.888 & $-$0.002 & 0.896 & $-$0.007 \\
\quad Basal ganglia, left$^\circ$ & T1 cohorts & 60 & -- & 0.699 & $-$0.009 & 0.762 & $-$0.006 & 0.737 & $-$0.011 \\
\quad Lateral ventricle, left$^\dagger$ & ABCD$\to$ADNI & 20 & -- & 0.693 & $-$0.164 & 0.821 & $-$0.037 & 0.759 & $-$0.071 \\
\quad White matter, left$^\dagger$ & ADNI$\to$ABCD & 20 & -- & 0.820 & $-$0.156 & 0.894 & 0.000 & 0.869 & $-$0.002 \\
\quad Thalamus, left$^\dagger$ & ABCD$\to$ADNI & 20 & -- & 0.748 & $-$0.004 & 0.819 & 0.000 & 0.777 & $-$0.002 \\
\quad White matter, left & ABCD$\to$ADNI & 20 & -- & 0.849 & $-$0.206 & 0.886 & $-$0.003 & 0.891 & $-$0.001 \\
\quad Brain mask$^\dagger$ & ABCD$\to$OASIS-3 & 20 & -- & 0.677 & $-$0.040 & 0.699 & $-$0.008 & 0.732 & $-$0.012 \\
\quad Brain mask & OASIS-3$\to$ABCD & 20 & -- & 0.617 & 0.000 & 0.622 & 0.000 & 0.662 & 0.000 \\
\quad Brain mask$^\dagger$ & ADNI$\to$OASIS-3 & 20 & -- & 0.723 & $-$0.010 & 0.744 & $+$0.005 & 0.741 & 0.000 \\
\midrule
\multicolumn{10}{l}{\emph{Modality translation (PSNR)}} \\
\quad T2$\to$T1ce$^\dagger$ & BraTS & 40 & 16.04 & 19.80 & $-$0.26 & 20.33 & $-$0.04 & 20.50 & $-$0.10 \\
\quad T1ce$\to$T2$^\dagger$ & BraTS & 40 & 16.22 & 20.42 & $-$0.14 & 20.73 & $+$0.05 & 20.74 & $+$0.05 \\
\quad FLAIR$\to$T1$^\circ$ & BraTS & 40 & 19.55 & 21.60 & $-$1.43 & 21.73 & $-$0.09 & 22.05 & $-$0.11 \\
\quad T2$\to$T1$^\dagger$ & IXI$\to$BraTS & 40 & 14.90 & 16.08 & $-$0.34 & 17.66 & $-$0.12 & 16.89 & $-$0.14 \\
\midrule
\multicolumn{10}{l}{\emph{Inpainting (PSNR inside the filled boxes)}} \\
\quad Inpainting, T1$^\dagger$ & BraTS & 40 & 3.87 & 16.42 & $-$3.13 & 18.80 & $-$0.18 & 18.33 & $-$0.45 \\
\quad Inpainting, T1ce & BraTS & 40 & 4.43 & 16.09 & $-$3.38 & 17.98 & $-$0.34 & 16.81 & $-$0.61 \\
\quad Inpainting, FLAIR$^\dagger$ & BraTS & 40 & 4.59 & 16.42 & $-$3.78 & 17.68 & $-$0.15 & 17.44 & $-$0.55 \\
\quad Inpainting, FLAIR, 5--6 boxes of side 12--20 & BraTS & 40 & 5.19 & 17.71 & $-$3.58 & 18.84 & $-$0.39 & 18.62 & $-$0.77 \\
\quad Inpainting, T1ce, one box of side 56--64 & BraTS & 40 & 3.55 & 14.95 & $-$2.49 & 16.66 & $-$0.14 & 15.12 & $-$0.29 \\
\quad Inpainting, T1ce, 5--6 boxes of side 12--20 & BraTS & 40 & 5.06 & 17.90 & $-$3.62 & 18.91 & $-$0.58 & 18.43 & $-$0.88 \\
\quad Inpainting, FLAIR, one box of side 56--64 & BraTS & 40 & 3.99 & 15.04 & $-$3.02 & 16.06 & $-$0.08 & 15.46 & $-$0.27 \\
\quad Inpainting, T2$^\dagger$ & IXI & 40 & 9.81 & 15.93 & $-$0.70 & 16.38 & $-$0.05 & 16.66 & $-$0.03 \\
\quad Inpainting, 5--6 boxes of side 12--20 & T1 cohorts & 60 & 6.50 & 18.84 & $-$2.13 & 19.34 & $+$0.03 & 19.32 & $-$0.02 \\
\quad Inpainting, one box of side 56--64 & T1 cohorts & 60 & 4.71 & 16.16 & $-$1.34 & 16.45 & $+$0.01 & 16.40 & $-$0.02 \\
\quad Inpainting, T2$^\circ$ & BraTS & 40 & 6.09 & 17.25 & $-$3.83 & 18.00 & $-$0.20 & 17.58 & $-$0.41 \\
\quad Inpainting, T1$^\circ$ & IXI & 40 & 8.41 & 16.35 & $-$1.44 & 16.23 & $+$0.10 & 16.70 & $-$0.10 \\
\quad Inpainting$^\dagger$ & IXI$\to$ABCD & 20 & 4.69 & 11.51 & $-$1.59 & 15.31 & $-$0.05 & 14.95 & $-$0.40 \\
\quad Inpainting & OASIS-3$\to$ABCD & 20 & 4.69 & 11.11 & $-$0.08 & 14.13 & $+$0.01 & 13.37 & $-$0.01 \\
\quad Inpainting$^\dagger$ & ABCD$\to$IXI & 40 & 8.41 & 9.61 & $+$1.63 & 12.10 & $+$0.24 & 12.06 & $+$0.08 \\
\quad Inpainting & OASIS-3$\to$ADNI & 20 & 4.76 & 11.85 & $-$0.08 & 14.16 & 0.00 & 13.57 & 0.00 \\
\quad Inpainting & ADNI$\to$IXI & 40 & 8.41 & 12.13 & $+$1.09 & 13.47 & $+$0.33 & 14.11 & $+$0.18 \\
\quad Inpainting$^\dagger$ & BraTS$\to$ABCD & 20 & 4.69 & 16.78 & $-$2.59 & 18.40 & $-$0.08 & 18.11 & $-$0.24 \\
\quad Inpainting & ADNI$\to$ABCD & 20 & 4.69 & 16.09 & $-$4.26 & 17.45 & $+$0.09 & 17.15 & $+$0.04 \\
\quad Inpainting$^\dagger$ & ABCD$\to$BraTS & 40 & 3.87 & 17.78 & $-$3.53 & 18.86 & $-$0.04 & 18.74 & $-$0.12 \\
\quad Inpainting & OASIS-3$\to$IXI & 40 & 8.41 & 12.47 & $+$0.14 & 13.42 & $+$0.03 & 13.05 & $+$0.04 \\
\quad Inpainting$^\circ$ & BraTS$\to$ADNI & 20 & 4.76 & 15.95 & $-$2.27 & 16.22 & $+$0.03 & 16.86 & $-$0.13 \\
\quad Inpainting$^\circ$ & ABCD$\to$ADNI & 20 & 4.76 & 16.54 & $-$2.45 & 16.58 & $+$0.13 & 17.29 & $-$0.06 \\
\midrule
\multicolumn{10}{l}{\emph{Bias correction (PSNR)}} \\
\quad Bias correction, $3^3$ grid, log-strength 0.4--0.8$^\dagger$ & T1 cohorts & 60 & 25.67 & 26.92 & $-$4.32 & 27.99 & $-$0.18 & 27.75 & $-$0.04 \\
\quad Bias correction, $3^3$ grid, log-strength 0.5--0.9 & T1 cohorts & 60 & 24.24 & 26.17 & $-$3.84 & 27.50 & $-$0.18 & 26.74 & $-$0.03 \\
\quad Bias correction, $3^3$ control grid$^\dagger$ & T1 cohorts & 60 & 27.67 & 27.57 & $-$4.84 & 28.42 & $-$0.21 & 28.41 & $-$0.08 \\
\quad Bias correction, $3{\times}4{\times}4$ grid & T1 cohorts & 60 & 25.84 & 27.04 & $-$4.25 & 27.99 & $-$0.16 & 27.22 & $-$0.09 \\
\quad Bias correction, $3^3$ control grid$^\dagger$ & BraTS$\to$ABCD & 20 & 25.86 & 26.05 & $-$6.58 & 26.83 & $-$1.31 & 27.19 & $-$2.17 \\
\quad Bias correction, $3^3$ control grid$^\dagger$ & ADNI$\to$ABCD & 20 & 25.86 & 26.98 & $-$6.45 & 28.17 & $-$0.20 & 27.72 & $-$0.03 \\
\quad Bias correction, $3^3$ control grid$^\dagger$ & ABCD$\to$BraTS & 40 & 25.52 & 25.57 & $-$2.49 & 26.39 & $+$0.05 & 26.30 & $-$0.30 \\
\quad Bias correction$^\circ$ & ABCD$\to$ADNI & 20 & 24.57 & 25.69 & $-$5.67 & 26.17 & $-$0.35 & 26.29 & $-$0.90 \\
\bottomrule\end{tabular}}
\end{table}
% ==== END tasks outside training ====